\pdfoutput=1
\documentclass[11pt]{article}

\usepackage[final]{acl}
\newcommand{\comment}[1]{}
\usepackage{times}
\usepackage{latexsym}
\usepackage{graphicx} % DO NOT CHANGE
\usepackage{amsmath}
\usepackage[T1]{fontenc}
\usepackage[utf8]{inputenc}

\usepackage{microtype}

\title{Controllable Response Generation for Assistive Use-cases}

\author{Shachi H Kumar\\ shachi.h.kumar@intel.com \And 
Hsuan Su \\hsuan.su@intel.com \And 
Ramesh Manuvinakurike \\ramesh.manuvinakurike@intel.com 
        \AND
        Saurav Sahay \\ saurav.sahay@intel.com  \And
        Lama Nachman \\ lama.nachman@intel.com }

\begin{document}
\maketitle
\begin{abstract}
Conversational agents have become an integral part of the general population for simple task enabling situations. However, these systems are yet to have any social impact on the diverse and minority population, for example, helping people with neurological disorders(eg. ALS), and people with speech, language and social communication disorders.%, sometimes with locked-in states where speaking or even typing is a challenge. 
Language model technology can play a huge role to help these users carry out daily communication and social interactions. To enable this population, we build a dialog system that can be controlled by users using cues/keywords. We build models that can suggest relevant cues in the dialog response context which is used to control response generation and can speed up communication. We also introduce a keyword-loss to lexically constrain the model output. We show both qualitatively and quantitatively that our models can effectively induce the keyword into the model’s response without degrading the quality of response. %We also show that providing fine-grained control helps in generating more diverse and contextually coherent responses. ? 
In the context of usage of such systems for people with degenerative disorders, we present human evaluation of our cue/keyword predictor and the controllable dialog system and show that our models perform significantly better than models without control. Our study shows that keyword-control on end-to-end response generation models is powerful and can enable and empower users with degenerative disorders to carry out their day-to-day communication. 
\end{abstract}

\section{Introduction}
Conversational agents, especially systems such as Alexa and Google Home, have become commodity items in people’s homes. Such systems have enabled carrying out one-shot tasks such as setting reminders, playing music and accessing information simpler for the general population. Besides the conversational agents that are popular with IoT and Mobile devices, we also have Personal Computer and Cloud based chatbots that are designed to perform certain goals or tasks, or to just engage in a casual conversation/chat with a user. The latter class of open-domain conversational agents have not yet seen widespread adoption besides mostly research exploration projects for developing conversational agents for long duration sustained and meaningful interactions \cite{ram2018conversational}. For example, Amazon launched the Alexa Prize challenge to develop Socialbots to converse coherently and engagingly with humans on popular topics such as Sports, Politics, Entertainment, Fashion and Technology for 20 minutes. Prior winning teams have used complex modular building blocks for intent detection, entity resolution, out of scope topic detection, dialog management strategies and a mix of template based and neural response generation methods \cite{Gabriel2020FurtherAI}.  

Large language models are also being developed today with end-to-end pre-training. Large-scale pre-training has attained significant performance gains across many tasks within NLP \cite{devlin-etal-2019-bert, Radford2018ImprovingLU}. Through self-supervised pre-training on large natural language corpora, these models gain generalized language understanding capabilities that transfer effectively to downstream tasks \cite{wang-etal-2018-glue}, including intent prediction \cite{Castellucci2019MultilingualID, Chen2019BERTFJ} and dialogue state tracking \cite{heck-etal-2020-trippy}. Open-domain chatbots are also being trained using generative language modeling objective of minimizing perplexity on next word prediction task using large conversational corpora and transformer based models. These models have demonstrated surprising generality, with models like DialoGPT \cite{zhang-etal-2020-dialogpt}, Meena \cite{Adiwardana2020TowardsAH} and Blender \cite{unknown} achieving response generation performance competitive with humans in certain settings. These improving systems still suffer from issues such as repeated responses, hallucinated facts, and lack of grounding and embodiment \cite{see-etal-2019-makes}. However, these models serve as a promising basis for further task specific fine-tuning and novel strategies to address issues such as repetitions, hallucinations and grounding.    

\begin{figure}
\begin{center}
    \includegraphics[width=\columnwidth]{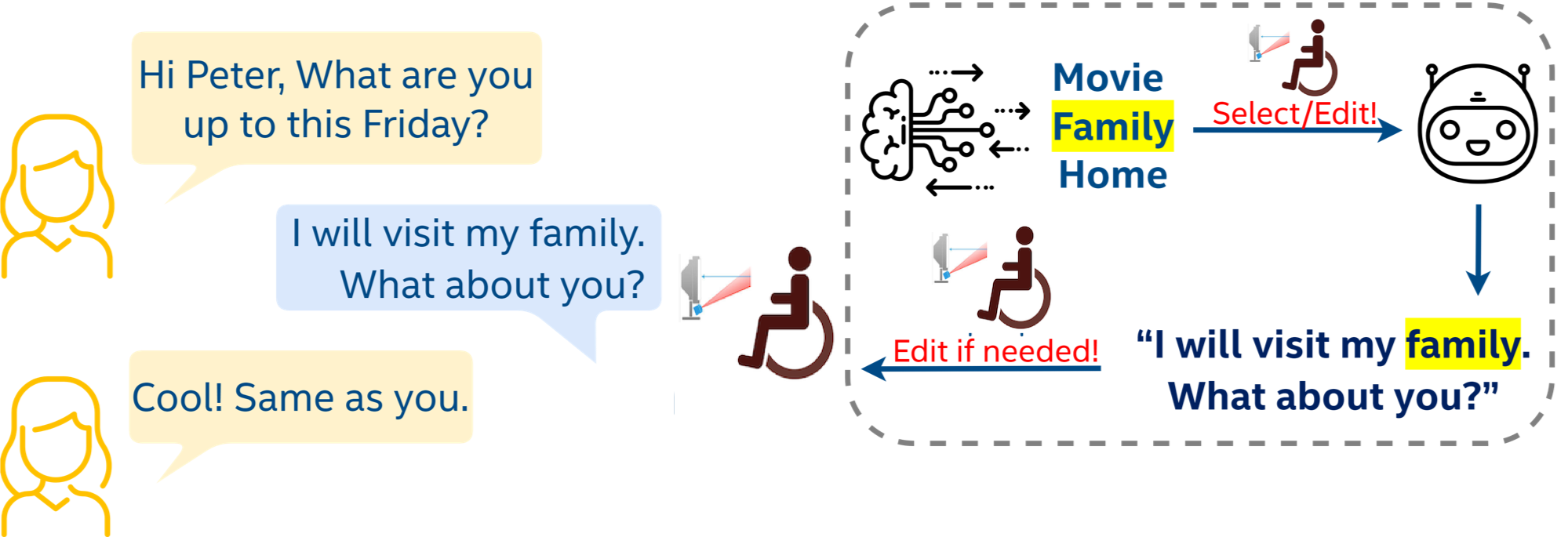}
\end{center}
\caption{A dialog system for an assistive use-case can listen to a conversation and provide diverse cues to the user. These cues, provide human control to the dialog system that can generate relevant responses that can further be edited. Such a system enables minimum effort and intervention from the user that is critical for people with disabilities.
}
\label{interaction_flow}
\end{figure}

Science and Engineering advances in Chatbots research make us hopeful for getting close to general purpose personalized AI assistants
%like J.A.R.V.I.S\footnote{https://ironman.fandom.com/wiki/J.A.R.V.I.S.} 
someday. Right now, with the availability of these pre-trained language enabling models, novel products and applications are emerging in niche domains \cite{bommasani2021opportunities} such as  Communication Systems (eg. email response completion \cite{10.1145/3292500.3330723}), Creativity Tools (story writing assistance \cite{roemmele2018automated,roemmele2021inspiration}), Human-AI collaboration for Software Engineering \cite{chen2021evaluating}, biosciences (protein structure prediction \cite{Rives622803} and several other emerging applications \cite{bommasani2021opportunities}.  
One such accessesibility application we are exploring is aimed towards leveraging language modeling technology to support minority group of people with certain disabilities \footnote{According to WHO, there are more than 1 Billion people with disabilities} 
to communicate with others effectively. For example, Amyotrophic Lateral Sclerosis (ALS) is a progressive, degenerative, neurological disorder that destroys the `motor neurons' that are responsible for enabling movement in arms, legs, chest, mouth and throat.
People with ALS lose their muscle movement, voice and the ability to carry out a normal day-to-day communication. It takes huge effort and time for these patients to communicate sentences character by character using various data input mechanisms available to them (gaze, fingers, muscle movements). 

We want to enable full and faster communication and provide interaction support tools for people with such disabilities by having an intelligent agent be their voice and content assistent. The system should use very limited user input (e.g. gaze, single muscle movement, facial gesture, etc) and suggest cues and cue-based responses that can be interactively chosen and edited for near real time social interactions. Today's response generation systems are very hard to use as-is for our usage requirements. The system for our usage needs to be context-aware, personalized, should enable minimal user-intervention and most importantly, be assistive and controllable. There has been some recent work on personalization, stylization and controllability of the generative language models, however, current research development tools are not suitable for our assistive usages. 

The system goal is to minimize the number of keystrokes input required for continued coherent interactions. Fast response generation with response cues, editing and auto complete features can dramatically reduce the silence gap in the conversation resulting from users slower keystroke by keystroke input. With these goals, we present a technique to control or guide a response generation model to generate a response that is relevant to the conversational context. 

Other possible application of this technology includes supporting people with other conditions such as Social Communication Disorder (SCD) \footnote{According to American Speech-Language-Hearing Association, pragmatic language impairment can exist in as much as 7.5\% of kindergarteners based on a sample study} or other pragmatic language impairments
such as difficulty in using language for social purposes, appropriately matching communication to the social context, following rules of the communication context (e.g., back and forth of conversation), and understanding nonliteral language (e.g., jokes, idioms, metaphors) \cite{doi:10.1177/1362361318822503}. 
Those with the diagnosis are impaired in processing implied sentences and indirect uses of language such as metaphors, humor, and aphorisms. They also display nonverbal communication problems along with verbal ones, such as greeting others according to context, waiting for turns in conversations, and modulating their behaviors according to context \cite{article_scd}. Our system provides cues and prototypical responses that can potentially be used by the community as aids in their interaction support and training. 

% \begin{figure}[h]
% \begin{center}
%     \includegraphics[scale=0.4]{images/user_image.png}

% \end{center}
% \caption{A user with disabilities communicating with a computer using eye-gaze, gestures and limited muscle movements.}
% \label{ACAT}
% \end{figure}

Our contributions in this work are, i) \textbf{Minority Group Application:} We bring forth a novel usage for open-domain chatbots/response generation systems, i.e., designing a reponse generation system that will represent users with communication disabilities and help them fulfill their day-to-day communication needs. ii) \textbf{Minimal user intervention:} We present a technique for fine-grained conversational controllable response generation using keywords/cues. We also build keyword/cue predictor models that further speed up communication time, and present human and automatic evaluation for these.  iii) \textbf{Keyword Loss:} We introduce a keyword loss to our training objective that further helps in incorporating soft lexical constraints in the form of keywords/similar words in the generated responses.
We show through automatic and human evaluation 
%based on several metrics including interaction time metrics, and show 
that our models are able to induce the keyword or its semantically similar meaning into the generated responses. 
\section{Related Work}

    \textbf{Assistive Technologies}: 
    %Studying Assistive Technologies for accessibility is important. 
    In recent times, AI augmented technologies have been developed that can help blind people better sense the visual world,
    %using novel egocentric vision devices
    offer real-time captioning and Sign Language interpretation for people with limited hearing, help augment the capabilities of people with limited mobility using new robotic systems, use brain computer interfaces for helping people communicate and help people with speech impairments better communicate. \cite{10.1145/2470654.2481291, 10.1145/3386296.3386298, 10.1145/3025453.3025814, ASL_Elakkiya, 9134708, ozawa2020word, Ramli_2020, Shor_2019}. 
    In ALS, individuals gradually loose their natural speech abilities due to the  reduction in speaking rates, rapid deteroriation of speaking ability and/or finger movement \cite{article_als1} and need Augmentative and Alternative Communication (AAC) strategies. AAC strategies support communication related to a large variety of issues, such as personal and medical care, social interaction and closeness, community involvement and employment, maintain emotional connection within families, provide patients with some autonomy and reducde caregiver burden \cite{10.3389/fneur.2018.00603}. Patients work with Speech Language Pathologists to learn to use low-tech to hi-tech AAC interventions to maintain Quality of Life after disease progression \cite{doi:10.3109/21678421.2015.1125499}. AAC interventions include speech generation \cite{article_als1}, eye-tracking tools \cite{gibbons2010functional} to BCI interfaces \cite{article_bci} for users with various degrees of locked-in states. Current communication systems use well-designed interfaces with inputs via eye-gaze or touch with some predictive text capability for word completion using simpler n-gram based language models \cite{verbally, therapy-box} and do not exploit the potential of using response generation technology using deep learning based language models. To the best of our knowledge, there aren't many research explorations for conversational technology based applications that exploit the latest language modeling techniques for people with MND such as ALS.
    \\
    \textbf{Controllable Generation}: Powerful text generation systems \citep{Radford2018ImprovingLU}, \citep{DBLP:conf/nips/BrownMRSKDNSSAA20} have emerged, however they are not controllable. \citep{DBLP:journals/corr/abs-1909-05858} pretrain a conditional transformer model with different types of control codes. \citep{DBLP:conf/emnlp/XuPSPFAC20} presents a keyword controlled technique for generating story endings. While the above involve modifying the language model itself, \citep{DBLP:conf/naacl/SeeRKW19} present post-processing techniques to control generated text. \citep{DBLP:conf/iclr/DathathriMLHFMY20} show an interesting plug-and-play architecture, where the base large language model is untouched, but one can introduce small attribute models to induce the control. \citep{DBLP:conf/emnlp/MadottoILDF20} extend the above plug-and-play architecture to dialog, and build attribute models for controls such as styles and topics. \citep{DBLP:journals/corr/abs-2009-10855} present techniques for controlling style in dialog and \citep{DBLP:journals/corr/abs-2008-09075} control response generation using semantic exemplars. However, all of these controllable attributes such as `topics', `sentiment' and `style' are too broad and not suitable for our use-case. We need fine-grained control that can enable minimal intervention from users.
   In the area of fine-grained controllable generation,
   \cite{xu2020generating} present a fine-grained guidance to the dialogue system to make the response meet user's expectation. \citep{ghazvininejad-etal-2017-hafez, see-etal-2019-makes} propose modifying the decoding procedure with several scoring functions to steer the bag-of-words on poetry generation. Both \cite{DBLP:conf/iclr/DathathriMLHFMY20} and \citep{ghazvininejad-etal-2017-hafez} present techniques for fine-grained generation, but the techniques are very time-consuming and require a lot of computational resources at decoding process which is not applicable to  real-time conversation scenarios.
   \\
    \textbf{Similarity-based Loss Function}: To induce the concept of keyword to the model, several works have focused on addressing the loss functions during model training. \cite{kovaleva-etal-2018-similarity} address the problem of representation learning and they successfully enhance the diversity and meaning in the generated sentence by using similarity-based losses. \cite{sha-2020-gradient} aims to lexically constrain the language generation, they propose to use the largest gradient between contrained word(keyword) and predictions to find the words that need to be updated. In our work, we aim to compute the loss across the entire sentence to guide keyword generation, rather than at word level. We hence try several variations to enable this. 
\section{Keyword and Response Modeling}

In this work, we modify the HuggingFace TransferTransfo model \citep{DBLP:journals/corr/abs-1901-08149} architecture initializing the decoder with the DialoGPT \citep{DBLP:conf/acl/ZhangSGCBGGLD20} weights. We incorporate fine-grained keyword-based control as model inputs and fine-tune the model on the DailyDialog \cite{li2017dailydialog} dataset with multi-task objective with an additional keyword based loss function. We enable keyword-control by 1) providing automatically generated keywords as auxillary input to the model and 2) by introducing a novel keyword-based loss that encourages the model to generate sentences containing the keyword or words semantically similar to the keyword. 

The model is similar to the Transformer based architecture from \cite{Radford2018ImprovingLU} that uses autoregressive and discriminative fine-tuning by optimizing a combination of two loss functions : 1) language modeling loss and 2) next-utterance classification loss. Given a context, the next utterance classification objective picks the next utterance among a set of candidates. We initialize this model with weights of DialoGPT, a large conversational response generation model and fine-tune on popular open domain conversation datasets with fine-grained control information. The DialoGPT model has the same architecture as GPT2 and is pretrained with millions of dialogs from Reddit conversations, making it more suitable for a dialog response generation task.

\subsection{\textbf{Keyword based Control}}
Given the conversation context, we enable fine-grained control over the responses generated by training the model with important cue words (we will refer to these as \textit{keywords} for simplicity) automatically generated from the response. 

\subsubsection{Keywords as context}
For a given conversation context, we incorporate keywords into the model by adding new keyword-specific-tokens, in addition to dialog-state/speaker tokens that represent speaker turns in the dialog. We further extend the dialog-state embeddings to add `keyword-state-embeddings' with special keyword separator token to indicate the positions of the keyword tokens. 
%For a given conversation context, we add dialog-state  embeddings  indicating  speaker  turns.  We  further  ex-tend  the  dialog-state  embeddings  to  add  ’keyword-state-embeddings’  to  indicate  the  positions  of  the  keyword  to-kens. 
%We append this keyword representation to the beginning of the dialog context that is input to the model. Figure \ref{modelarch} shows the combined embeddings fed into the model consisting of position, token and dialog state embeddings whhere the dialog state is comprised of keyword-state and speaker-state. 
% \begin{figure}[htbp]
%   \centering
%   \includegraphics{LaTeX/images/process.png}
%   \caption{svg image}
% \end{figure}
\subsubsection{Keyword-based loss functions} We propose keyword-based loss functions that encourage the occurence of the input keyword(s) in the generated sentence. We introduce variations to this loss function to enable the generation of semantically similar word to the input keyword as well as incorporate multiple-inputs (where the goal is to induce multiple keywords in the response) as control to the model. With addition of this loss, the overall loss of the model is a combination of : language model loss $L_{m}$, next sentence prediction loss $L_{n}$ and keyword loss $L_{k}$,
\begin{equation} \label{eq1}
\begin{split}
  \text{Overall Loss}, L = \alpha L_{m}+ \beta L_{n} + \gamma L_{k}
\end{split}
\end{equation}
where $\alpha$, $\beta$ and the $\gamma$ are the hyper-parameters. $\alpha$ and $\beta$ are set to 1 as in the original TransferTransfo model.

    \paragraph{\textbf{Keyword Loss:}}
    In order to encourage the generation of the cue/keyword in a sentence, our goal is to maximize the similarity between the keyword and one of the generated words (at some output position). So to generate a keyword $kw$ at some output timestep, we derive the probability distribution from the generated logits at every timestep i=1 to T, and compute the negative log of the probability of the $kw$ at each step. We then take the minimum of these scores across the generated sentence. Hence, the loss over the sentence w.r.t the keyword K is equal to, 
    \begin{equation} \label{eq2}
\begin{split}
\; L_{k} = \min_{i=1}^T(-\log p_i(kw)), \\
%  {\color{red}  \text{Keyword\ Loss}, \; L_{k} = \min_{t=1}^T(CE(K, x_t)), } \\
%  {\color{red} where \; CE(K, x) = - \sum_i^V  K_i*\log(\widehat{x_i})}
\end{split}
\end{equation}
   
    \paragraph{\textbf{Keyword Loss with similar words:}} We incorporate embedding-based similarity scores into the keyword loss computation as shown in equation \ref{eq3} in order to encourage generation of not just the keywords, but also semantically similar words in the sentence. Let $pool = kw \cup sim\_words(kw) $. The KeywordLoss $L_k$, 
    % \begin{itemize}
        % \item     K is a vector such that:
%   K_i = \begin{align}
%     \begin{equation} \label{eq1}
\begin{equation}\label{eq3}
\begin{split}
L_{k} = sim(k, kw)
\min_{i=1}^T(-\log p_i(k)),\\
where \; k = \arg\min_{x \in pool}(\min_{i=1}^T(-\log p_i(x)))
\end{split}
\end{equation}

% \left\{\begin{matrix}
% 1, & i=kw\\ 
% sim(kw,word_i), & i \; \epsilon \; top_N sim\_words(kw) \\ 
%  0, & otherwise 
% \end{matrix}\right.

% \end{equation}\end{align}
% \begin{figure}[h!]
% \includegraphics[scale=0.4]{LaTeX/images/simscoreLoss.png}
% \caption{Placeholder image for equation}
% \label{simscoreloss}
% \end{figure}
% \begin{equation}
%     K_i = \left\{\begin{matrix}
%     1 & i = kw \\
%     sim(kw, word_i) & word_i\in top_N(Similar(kw))\\ 
%     0 & otherwise 
% \end{matrix}\right.
% \end{equation}

% \item     K is a vector such that:

% \begin{figure}[h!]
% \includegraphics[scale=0.4]{LaTeX/images/simscore1.png}
% \caption{Placeholder image for equation}
% \label{simscore1}
% \end{figure}
% \begin{equation}
%     K_i = \left\{\begin{matrix}
%     1 & i = top_N(Similar(kw)) \\
%     0 & otherwise 
% \end{matrix}\right.
% \end{equation}
%     \begin{equation} \label{eq1}
% \begin{split}
% \begin{align}
%     K_i = 
% \left\{\begin{matrix}
%   1, &  i = kw \;or \;i \epsilon \;topN\; sim\_words(kw)}\\ 
%  0, & otherwise 
%  \end{matrix}\right.
% \end{align}
% \end{split}
% \end{equation}
% \end{itemize}

    \paragraph{\textbf{Keyword Loss with multiple inputs}}: Lets consider $k_{1}, k_{2}... k_{N}$ as the $N$ multiple control inputs. In this scenario, it is desirable that the generated output contains all of the keywords (or words similar to the keywords). To enable this, we minimize the negative log probability for each keyword, $k_j$,  across the entire sentence and add these scores as the total loss. 
% \begin{equation} \label{eq1}
% \begin{split}
%   L_{multi-kw} = (\sum_i^N{(min_t(\sum_j^V - K_{ij}log(p_j))))}
% \end{split}
% \end{equation}
\begin{equation}
 L_{k} = \sum_{j=1}^{N} \min_{i=1}^T(-\log p_i(k_j))
\end{equation}

\comment{
\begin{figure*}
\includegraphics[width=\textwidth]{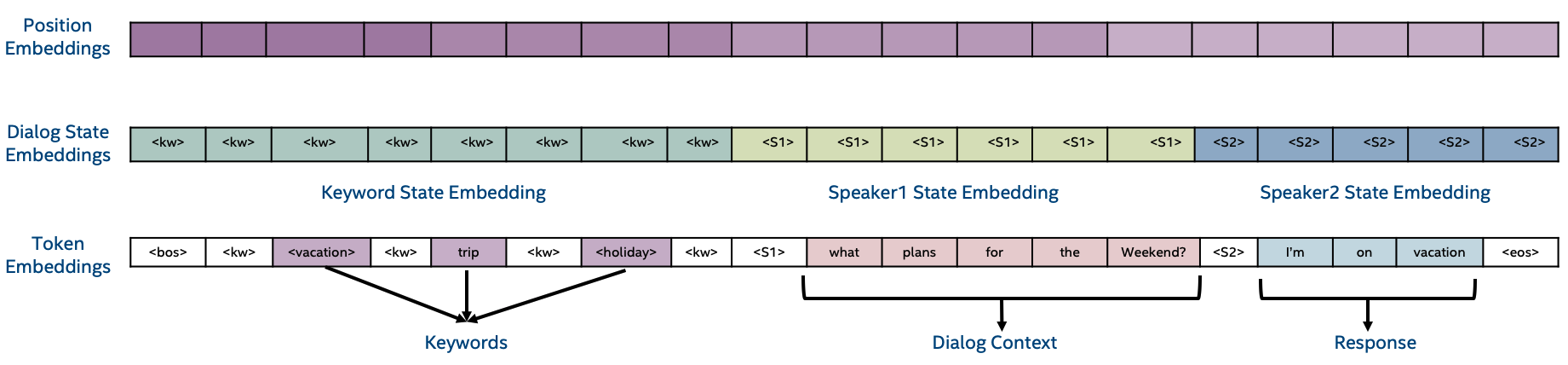}
\caption{Inputs to the model is a combination of position, dialog state and token embeddings. Fine-grained control in the form of keywords is achieved by defining keyword separator tags and keyword-state-embeddings added to the input.}
\label{modelarch}
\end{figure*}
}
\subsection{Keyword Generation}
\paragraph{Training:} We extract `key' terms from the dataset to fine-tune the model - this data is generated automatically, hence enabling end-to-end automatic pipeline, without the need for any other additional data collection or labeling efforts. Given a conversation context and a response output, keywords are extracted from the response utterance and incorporated into the model. We use keyBERT \citep{grootendorst2020keybert} to extract meaningful keywords from the responses. This technique uses BERT-embeddings and cosine similarity to find the sub-phrases in a document that are most similar to the document itself. We generate top-3 1-gram keywords (rather than key-phrases) for each dialog response, as 1-gram inputs are most suitable for our use-case. We use both single keywords and multiple keywords as inputs to the models in our experiments. 
\paragraph{Inference:} During inference, especially in our use-case where we need to minimize user intervention, the user would greatly benefit from automatic keyword suggestions, rather than having to type it out each time. Hence we build keyword prediction models, given conversation context as input. We build two models for keyword prediction: \\
{\textbf{1) Extractive keyword predictor:}}
Given a conversation context, we use DialoGPT\cite{zhang2019dialogpt} with diverse beam search\cite{Vijayakumar_Cogswell_Selvaraju_Sun_Lee_Crandall_Batra_2018} to generate multiple responses (we use 10 beams, 2 groups and diversity\_penalty of 5.5). We then use keyBERT\cite{grootendorst2020keybert} to extract keywords from the beam outputs and present these as keyword suggestions. \\
{\textbf{2) Generative keyword predictor:}} We fine-tune a large pretrained language model, GPT2, to generate keywords for a given context, and present these as suggestions. We use the training and validation dataset from DailyDialog to build the keyword predictor. This generative predictor is trained to predict multiple keywords for a given context. For evaluation of these models, we use the top keyword prediction. We further use diverse beam search(same configuration as above) and generate multiple keyword suggestions.

\section{Methods}
We compare our keyword-models (without the keyword loss but containing the keywords as part of dialog context) and keyword-loss-based (with additional keyword loss) models on the DailyDialog dataset  \citep{DBLP:conf/ijcnlp/LiSSLCN17} %and the PersonaChat dataset \citep{DBLP:conf/acl/KielaWZDUS18}. 
% redundant We use the TransfeTransfo model with DialoGPT as our backbone architecture. % and compare the various fine-tuned models with and without embedded keywords. It is to be noted that our technique for Keyword control and expansion for K-CRG and S-CRG can be used on any other SOTA dialog model.  
We initialize the network with weights of DialoGPT `medium' model with 345M parameters. We also use 2 candidates for the next utterance prediction task. We use language modeling and multiclass-classification coefficients of 1. We use a batch\_size of 64 for training, nucleus sampling for generation with top\_p set to 0.9. We fine-tune the model for 3 epochs. 
We run experiments on 5 main classes of models:
i) No-keyword model (\textbf{$no\_kw$} henceforth): Trained without any keyword information 
ii) Keyword-context ($kw\_context$): Trained with keyword as auxillary input + dialog context
iii) Keyword-loss ($kw\_loss$): Incorporates keyword loss + keyword as auxillary information.
iv) Keyword sim-loss ($kw\_sim\_loss$): Incorporate similar words (embedding-based techniques such as Glove \citep{pennington2014glove} ($kw\_sim\_loss\_glove$) or wordnet-based ($kw\_sim\_loss\_wordnet$) similarity for loss computation . We experiment with 2 variations, one using the similarity score, and the other using 1.
v) Multiple-keyword-loss ($multi\_kw\_loss$): Incorporate multiple keywords into the input as well as into the loss computation.

\subsection{Datasets}
We use the Dailydialog dataset \citep{DBLP:conf/ijcnlp/LiSSLCN17}, which consists of 13,118 daily conversations involving various topics such as tourism, culture, and education among others. The validation and test sets
have 1000 conversations each. For the conversation context, we consider a maximum of upto 5 previous utterances as history for generation. We use the test set, consisting of 6740 context-response pairs, to evaluate our models which will be discussed in the results section. DailyDialog dataset contains daily life communication with the goals of exchanging ideas and information and enhancing social bonding. This dataset contains suitable interactions for building applications to support social communication and daily life interactions and serves as a starting point for AAC applications. 
%We also finetune and test our models on PersonaChat dataset \citep{DBLP:conf/acl/KielaWZDUS18}. It contains 8939 dialogues for training, 1000 for validation, and 968 for testing. We do not use the persona information in this work. 

%\subsection{Model Evaluation} 
\subsection{Automatic Evaluation} We use several automatic metrics to compute the performance of our models. Given the well-discussed fact that word-overlap based metrics do not agree well with human judgment, we utilize other learning based and embedding-based metrics to evaluate the generated response with the reference ground truth.

\subsubsection{{\textbf{Metrics for Evaluating Keyword Predictor Models}}}
The keyword predictor model should be able to generate diverse keywords to present varied options for users to choose from. We evalute the extractive and generative models based on the diversity of keywords generated. We use averaged cosine similarity between generated keywords as a measure of diversity-lower the similarity, higher the diversity.
%- we use based on the keywords they generate. Our models need to generate diverse keywords so that the users can effectively disambiguate and make a choice. With this in mind, we compute the average cosine similarity scores between the glove embedding representations of the keywords generated from the predictor model. 
We hypothesize that meaningful keywords will result in generation of meaningful and context-relevant responses. Hence, we compute `human-like' and coherence scores for the generated responses using DialogRPT \citep{gao2020dialogrpt}, a model trained to predict human feedback dialogue responses. 
\subsubsection{{\textbf{Metrics for Evaluating Controllable Response Generation Model}}}
\paragraph{\textbf{Keyword Insertion Accuracy(KIA):}}
The main goal of this work is to provide fine-grained control to the user to have the model induce a keyword (or a word with a similar meaning) in the response. %Hence, we are able to objectively evaluate the insertion against the reference ground truth. %Given an input word, the model need not necessarily always generate a response `containing' the word, but it could also contain words with semantically similar meanings. 
% The main goal of this work is to provide fine-grained control to the user to have the model generate a response that is relevant to the given input keyword. Since our intention is to induce the keyword (or a word with a similar meaning) in the response, we are able to objectively evaluate the insertion against the reference ground truth. Given an input word, the model need not necessarily always generate a response `containing' the word, but it could also contain words with semantically similar meanings. %One way  to automatically evaluate the effectiveness of the models is to identify if the input word or a word that is similar, is a part of the generated sentence or not. 
With this in mind, we compute the keyword-insertion accuracy to evaluate the controllable response generation models. %in two ways 1) We compute the accuracy of exact keyword insertion 2) we also compute accuracies of insertion of words containing similar meaning into the generated response. We use embedding-based cosine similarity metrics and heuristically use a threshold 0.7 to compute the accuracies. %Automatic Threshold hyperparameter tuning to improve the generation is to be explored as part of our future work. 

\paragraph {\textbf{Similarity Based Metrics:}}
For the metrics in this section, we consider the conversation context, generated response and the ground truth for evaluation. Because we intend to generate responses based on keywords, computing measures of similarity between the generated response and ground truth response (in the learnt embedding space) gives a good assesment for the model performance. We use {BLEURT}, {BERTScore} \citep{DBLP:conf/iclr/ZhangKWWA20} \citep{DBLP:conf/acl/SellamDP20}, {Sentence-BERT} \citep{DBLP:conf/emnlp/ReimersG19} to compute similarity between generated response and ground truth.

% \begin{itemize}
%     \item :A learned evaluation metric based on BERT and shown to be more correlated with human judgement. 
%     \item :State-of-the-art embedding-based evaluation method based on BERT. 
%     \item : Obtains rich and semantically meaningful sentence embeddings. We compute the cosine similarity between the S-BERT embeddings of the response and ground truth.
% \end{itemize}

\paragraph{\textbf{Response Quality Metrics:}} 
Given the one-to-many nature of open domain dialog, although the above metrics provide a good sense of the generated response, evaluating the quality of response is essential. we only focus on turn level response quality aspects such as fluency, context coherence and diversity. We perform language model based evaluation for fluency and context-coherence and n-gram based diversity evaluation, % \footnote{https://github.com/alexzhou907/dialogue\_evaluation}, based on \citep{DBLP:conf/acl/PangNHZLT20}. 
We also measure the perpelexity (PPL) by employing a ptretrained GPT-2 "medium" model.
%\subsubsubsection{\textbf{Response Length}} As it has been shown in several works including \citep{DBLP:conf/naacl/SeeRKW19},\citep{DBLP:conf/naacl/LiGBGD16} a common issue of large language models is they tend to favor or generate generic responses. We compute the length of the generated sentences of the various models with this following hypothesis - the longer the responses, the less generic they tend to be.

\comment {
\begin{table}
\centering
\begin{tabular}{lrl}
\hline \textbf{Model} & \textbf{DailyDialog} & \textbf{PersonaChat}\\ \hline
NoKeyword & 7.68 & 8.79\\
K-CRG & 9.29 & 11.48\\
Syn\_W-CRG  & 8.57 & 9.96 \\
Syn\_G-CRG & 9.13 & 10.55\\
\hline
Ground Truth & 14.067 & 11.96\\
\hline
\end{tabular}
\caption{\label{meanLen} {\color{red} This table will be commented out}Mean Length of Generated Sentences from Different Models (on DailyDialog dataset)}
\end{table}
}

\subsection{Human Evaluation}
% Since, we focus on evaluating the system for assistive system use case,
%, standard approaches involving the users rating the responses through questionnaires is not sufficient \textcolor{red}{(citations mesuring responses)}. 
%  we obtain the ratings of the responses and also measure the interaction time  while generating the response in a conversation. %We develop an evaluation pipeline with focus on measuring the overall time saved along with the human ratings of the responses. 

We perform human evaluation via Amazon Mechanical Turk to evaluate the keyword predictor models and controllable response generation models. %We also adopt metrics such as %keystroke savings rate \cite{trnka2008evaluating}, which is typically used in measuring the word prediction tasks, and 
%interaction time \cite{devault2011incremental}, which is typically used in measuring the dialogue systems response latency. These are crucial to evaluate the performance of our models for the assistive use-case as they're typically more indicative of effectiveness of the models usefulness. 
Towards measuring this we perform evaluations in 3 separate crowd-tasks. 

\paragraph{Task1: Collecting response for automatic and human-entered keywords}
We present a conversation context and keywords (from the extractive and generative keyword prediction models) to the turkers and ask them to come up with possible responses relevant to these keywords. The turkers are also asked to answer a question about the relevance of each keyword to the provided context. This information is also used to evaluate the keyword predictor models. To represent human-control in our analysis, the turkers are also asked to enter keywords of their choice, along with the corresponding responses. %This step is essential to collect human-generated control data. %Data collected in this step is used in Task2.
\paragraph{Task2: Overall system interaction and metrics:} 
We use the data obtained in Task1 (\textbf{keywords}:human and model-generated, and the \textbf{responses} (we treat these as ground truth human responses)). In the interaction flow, the user reads the conversation context, picks a keyword that he/she wants to respond with - which brings up a human response and a model response ($kw\_loss$ model). The user can either pick one of the presented responses or edit them partially or type a new response altogether. We present the user with a set of questions based to enables us to understand the reason for the choice made by the user. We analyse if the users tend to choose a model or a keyword response and also compute the word error rates (WER) for the edits corresponding to the human and model responses. %In this interaction, we measure the interaction time, i.e., time taken to make the keyword and response choice altogether. 
% obtained by integrating our keyword predictor and control-based response generation model. % When the user edits a particular response, we also compute the edit distance of the response from the choice. 

\paragraph{Task3: Human Evaluation of controllable response generation models:}
To perform human evaluation of the quality of the responses, %with aspects such as fluency, context relevance, generic/specificity of the responses, we perform human evaluation using crowd-sourcing. 
we randomly pick 100 dialog contexts and present the context along with the keyword and pairs of responses from the models and ask 3 annotators to rate the responses based on the following criteria: 1) Fluency: how natural and fluent the responses are, 2) Generic: are the responses too generic given the dialog context?, 3) Context relevance: how relevant and coherent is a response to a given dialog context, 4) Keyword relevance: how relevant is a response to the input keyword? 
% \begin{itemize}
% \item \textbf{Fluency}: Are the response natural and fluent?
% \item \textbf{Generic}: Are the responses generic given the conversation context?
% \item \textbf{Context Relevance}: Are the responses relevant and coherent given the conversation? 
% %Does the response follow the conversation context? 
% \item \textbf{Keyword Relevance}: Are the responses relevant to the keyword provided? 
% \end{itemize}

We present pairs of responses from models A and B and provide 4 options for for each of the above criteria: A better than B, B better than A, Both and, Neither. 
%\textcolor{red}{DO WE NEED THIS? \textbf{We also ask the annotators to write a 1-2 line summary of the conversation to be sure that they are rating the models effectively.}} 
We evaluate the pairs, $no\_keyword$ vs $kw\_context$, $no\_keyword$ vs $kw\_loss$ and $kw\_context$ vs $kw\_loss$. We compute the scores using a majority vote across 3 annotators.  
%{\color{red}Mention about how much we pay, etc}

\section{Results}

\begin{table}[tp!]
\resizebox{\linewidth}{!}{
\begin{tabular}{|c|c|c|c|}
\hline
Kw Predictor & Coherence      & Human-like     & Diversity$\downarrow$      \\ \hline
Generative           & \textbf{0.903} & \textbf{0.641} & \textbf{0.227} \\ \hline
Extractive     & 0.891          & 0.595          & 0.265          \\ \hline
\end{tabular}
}
\caption{\label{keywordpred} Evaluation of keyword predictor models.}
\end{table}

In this section we present the results of human and automatic evaluation on the performance of our keyword-based controllable response generation models and the keyword predictor models.

\begin{table*}[tp!]
\centering
\resizebox{\textwidth}{!}{
\begin{tabular}{|lccclcccc|}
\hline
\multicolumn{1}{|l|}{\textbf{}}                         & \multicolumn{1}{c|}{KIA}            & \multicolumn{1}{l|}{Similarity}     & \multicolumn{1}{c|}{BLEURT}           & \multicolumn{1}{l|}{BERT Score}                 & \multicolumn{1}{l|}{Context}        & \multicolumn{1}{l|}{Diversity}      & \multicolumn{1}{l|}{Fluency}        & \multicolumn{1}{l|}{PPL$\downarrow$} \\ \hline

\multicolumn{9}{c|}{\textbf{Single Keyword}}  \\
 
\cline{1-9}
% &  \multicolumn{1}{c|}{}          & \multicolumn{1}{c|}{}          & \multicolumn{1}{c|}{}          & \multicolumn{1}{l|}{}          & \multicolumn{1}{c|}{}          & \multicolumn{1}{c|}{}          & \multicolumn{1}{c|}{}          &   \\ 

\multicolumn{1}{|l|}{no\_kw}                     & \multicolumn{1}{c|}{0.083}          & \multicolumn{1}{c|}{0.271}          & \multicolumn{1}{c|}{-1.035}          & \multicolumn{1}{l|}{0.868/0.836/0.851}          & \multicolumn{1}{c|}{0.541}          & \multicolumn{1}{c|}{1.592}          & \multicolumn{1}{c|}{\textbf{0.407}} & \textbf{39.098}          \\ \hline
\multicolumn{1}{|l|}{kw\_context}              & \multicolumn{1}{c|}{0.672}          & \multicolumn{1}{c|}{0.539}          & \multicolumn{1}{c|}{-0.607}          & \multicolumn{1}{l|}{0.844/0.853/0.868}          & \multicolumn{1}{c|}{0.568}          & \multicolumn{1}{c|}{\textbf{1.789}} & \multicolumn{1}{c|}{0.403}          & {41.752}                   \\
\multicolumn{1}{|l|}{kw\_loss}                 & \multicolumn{1}{c|}{\textbf{0.694}} & \multicolumn{1}{c|}{\textbf{0.542}} & \multicolumn{1}{c|}{-0.609}          & \multicolumn{1}{l|}{\textbf{0.885/0.852/0.868}} & \multicolumn{1}{c|}{0.579}          & \multicolumn{1}{c|}{1.726}          & \multicolumn{1}{c|}{\textbf{0.407}} & 43.115                   \\ \hline
%kw\_sim\_loss                                  & \multicolumn{1}{l}{}                & \multicolumn{1}{l}{}                & \multicolumn{1}{l}{}                 &                                                 & \multicolumn{1}{l}{}                & \multicolumn{1}{l}{}                & \multicolumn{1}{l}{}                & \multicolumn{1}{l|}{}    \\
\multicolumn{1}{|l|}{kw\_sim\_loss\_glove-1}   & \multicolumn{1}{c|}{0.684}          & \multicolumn{1}{c|}{0.541}          & \multicolumn{1}{c|}{\textbf{-0.606}} & \multicolumn{1}{l|}{0.884/0.852/0.868}          & \multicolumn{1}{c|}{\textbf{0.585}}          & \multicolumn{1}{c|}{1.729}          & \multicolumn{1}{c|}{0.405}          & 42.544                   \\
\multicolumn{1}{|l|}{kw\_sim\_loss\_wordnet-1} & \multicolumn{1}{c|}{0.686}          & \multicolumn{1}{c|}{0.540}          & \multicolumn{1}{c|}{-0.615}          & \multicolumn{1}{l|}{0.884/0.852/0.868}          & \multicolumn{1}{c|}{0.581} & \multicolumn{1}{c|}{1.726}          & \multicolumn{1}{c|}{0.403}          & 42.606                   \\ \hline
\multicolumn{1}{|l|}{kw\_sim\_loss\_glove}     & \multicolumn{1}{c|}{0.680}          & \multicolumn{1}{c|}{0.543}          & \multicolumn{1}{c|}{-0.610}          & \multicolumn{1}{l|}{0.885/0.852/0.868}          & \multicolumn{1}{c|}{0.570}          & \multicolumn{1}{c|}{1.741}          & \multicolumn{1}{c|}{0.403}          & 42.362                   \\
\multicolumn{1}{|l|}{kw\_sim\_loss\_wordnet}   & \multicolumn{1}{c|}{0.672}          & \multicolumn{1}{c|}{0.541}          & \multicolumn{1}{c|}{-0.606}          & \multicolumn{1}{l|}{0.884/0.852/0.867}          & \multicolumn{1}{c|}{0.576}          & \multicolumn{1}{c|}{1.733}          & \multicolumn{1}{c|}{0.403}          & 42.301                   \\ \hline

\multicolumn{9}{c|}{\textbf{Multiple Keywords}}  \\
\cline{1-9}

% \multicolumn{1}{|l|}{\textbf{Multiple Keywords}}   & \multicolumn{1}{c|}{}          & \multicolumn{1}{c|}{}          & \multicolumn{1}{c|}{}          & \multicolumn{1}{l|}{}          & \multicolumn{1}{c|}{}          & \multicolumn{1}{c|}{}          & \multicolumn{1}{c|}{}          &   \\

\multicolumn{1}{|l|}{no\_kw} & \multicolumn{1}{c|}{0.041} %         & \multicolumn{1}{c|}{0.070} 
& \multicolumn{1}{c|}{0.271}          & \multicolumn{1}{c|}{-1.035}          & \multicolumn{1}{l|}{0.868/0.836/0.851}          & \multicolumn{1}{c|}{0.541}          & \multicolumn{1}{c|}{1.592}          & \multicolumn{1}{c|}{0.407}          & \textbf{39.098}                   \\ \hline
\multicolumn{1}{|l|}{kw\_context}      & \multicolumn{1}{c|}{0.293}      %    & \multicolumn{1}{c|}{0.122}      
& \multicolumn{1}{c|}{{0.607}} & \multicolumn{1}{c|}{\textbf{-0.499}} & \multicolumn{1}{l|}{\textbf{0.895/0.857/0.875}} & \multicolumn{1}{c|}{0.489}          & \multicolumn{1}{c|}{\textbf{1.396}} & \multicolumn{1}{c|}{0.399}          & {75.300}          \\
\multicolumn{1}{|l|}{kw\_loss}          & \multicolumn{1}{c|}{0.300}      %    & \multicolumn{1}{c|}{0.125}      
& \multicolumn{1}{c|}{0.604}          & \multicolumn{1}{c|}{-0.524}          & \multicolumn{1}{l|}{0.894/0.856/0.874}          & \multicolumn{1}{c|}{\textbf{0.492}} & \multicolumn{1}{c|}{1.354}          & \multicolumn{1}{c|}{0.412}          & 83.971                   \\ \hline
%kw\_sim\_loss                     & \multicolumn{1}{l}{}                & \multicolumn{1}{l}{}                & \multicolumn{1}{l}{}                & \multicolumn{1}{l}{}                 &                                                 & \multicolumn{1}{l}{}                & \multicolumn{1}{l}{}                & \multicolumn{1}{l}{}                & \multicolumn{1}{l|}{}   \\
\multicolumn{1}{|l|}{kw\_sim\_loss\_glove-1} & \multicolumn{1}{c|}{\textbf{0.302}} 
%& \multicolumn{1}{c|}{0.127}          
& \multicolumn{1}{c|}{\textbf{0.610}}          & \multicolumn{1}{c|}{-0.535}          & \multicolumn{1}{l|}{\textbf{0.895/0.857/0.875}} & \multicolumn{1}{c|}{0.487}          & \multicolumn{1}{c|}{1.366}          & \multicolumn{1}{c|}{0.416}          & 84.367                   \\
\hline

\end{tabular}}
\caption{\label{keywordModelPerf}Performance of the various controllable models for single and multi-keyword inputs ($\gamma$ = 0.005). Label "-1" indicates that we set $sim(k, kw)=1$ in equation \ref{eq3}.}
\end{table*}

\begin{figure}[h]
    \centering
    \includegraphics[width=\columnwidth]{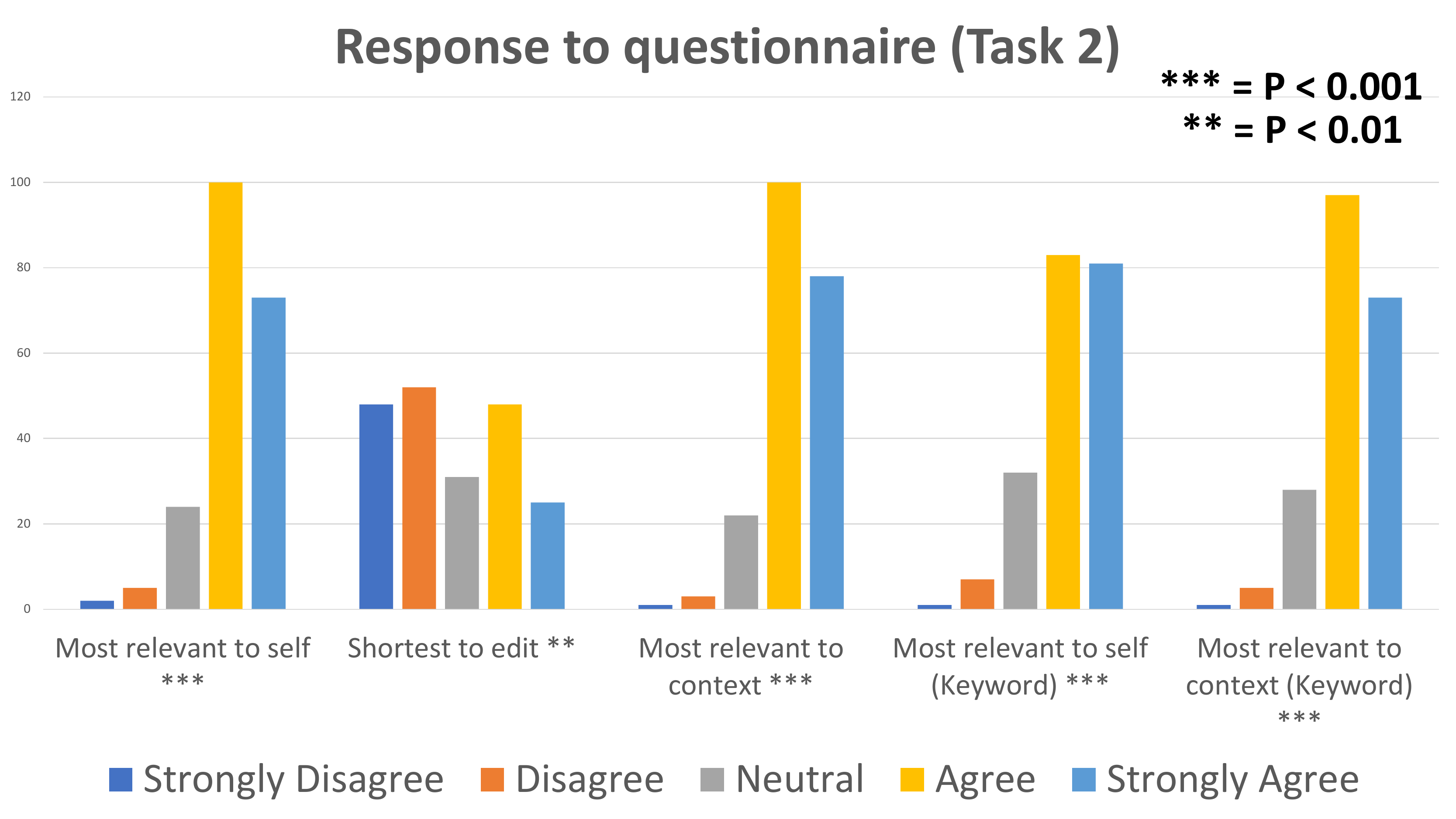}
    \caption{Shows the responses to the questionnaire. We observe that users signifiantly felt that they chose the keyword \& response as they were most relevant to what was on their mind. We also observe that the users disagree that they chose a response because of it's length. 
    (One-Sample Wilcoxon Signed Rank Test (mu=0)). }
    \label{fig:human_task2_eval}
\end{figure}
 
 \begin{figure}[h]
    \centering
    \includegraphics[width=\columnwidth]{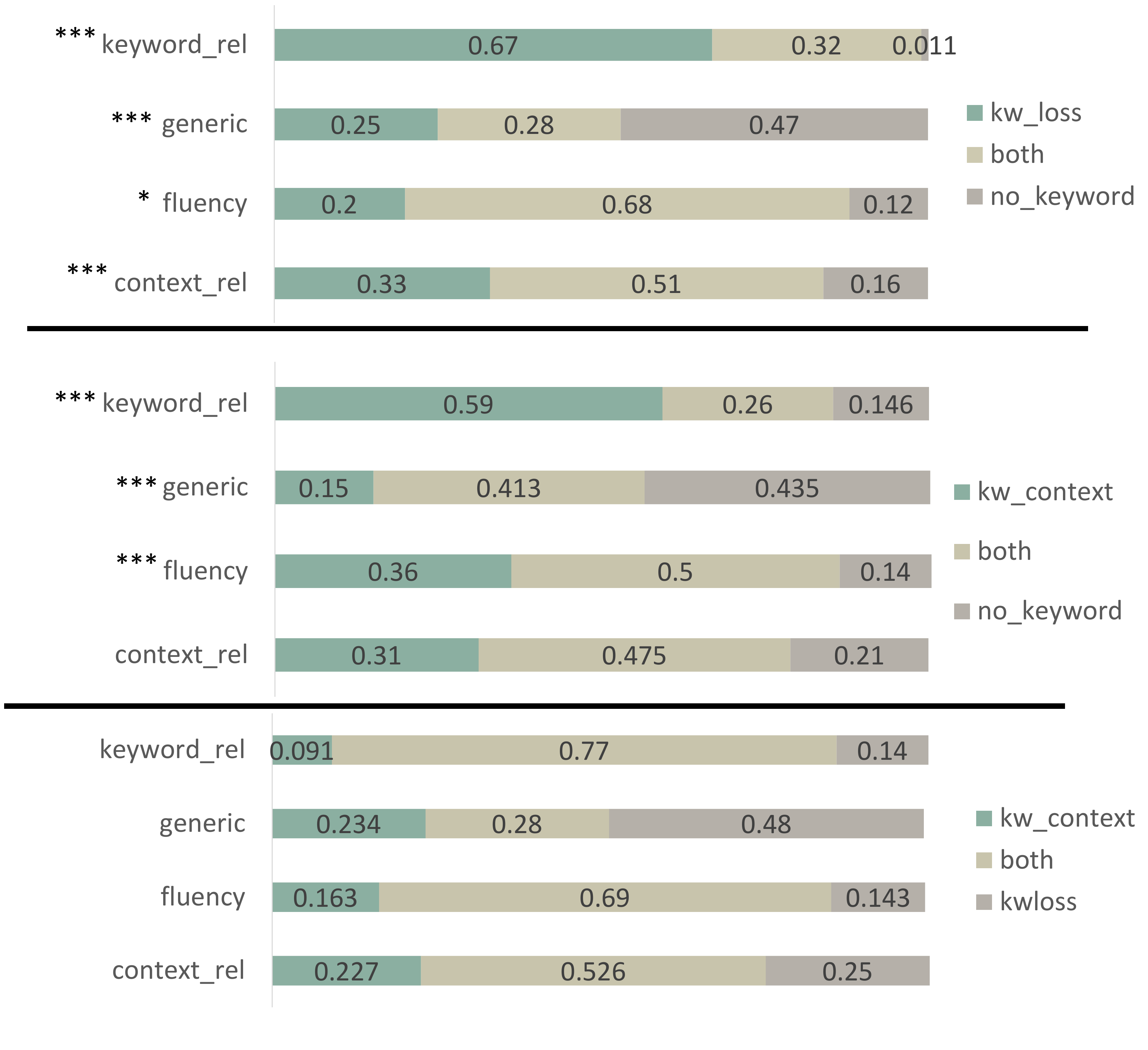}
    \caption{Results from human evaluation. (One-Sample Wilcoxon Signed Rank Test (mu=0) for the statistical tests.*** p\textless 0.001, ** p\textless 0.01, * p\textless 0.05.) }
    \label{fig:human_task3_eval}
\end{figure}

\subsection{Automatic Evaluation Results}
We present the evaluation results on the similarity and response-quality metrics computed on the conversation context, generated response and the reference response, on the DailyDialog testset.

\paragraph{Keyword Predictor Models:} Table \ref{keywordpred} shows the performance of the keyword predictor models based on diversity and the keyword-success-rate (coherence and human-like scores). The diversity of the generated keywords is measured using average cosine-sim computed between pairs of keywords generated - lower score indicates higher diversity.
We can observe that the generative keyword predictor tends to generate more diverse keywords (lower similarity score), which is very important in our use-case. The responses generated by choosing the keywords from the generative predictor are more coherent and human-like.

\paragraph{Cue/Keyword controlled models:} 
%\paragraph{Effect of $\gamma$:}We experiment with the keyword-loss models with various values of $\gamma$ ranging between 0 and 1. We see the best performance when $\gamma$=0.005 (more details in the Appendix section). Henceforth, unless otherwise mentioned, we will use $\gamma$=0.005 for all our experiments. 
We experiment the keyword-loss models with various values of $\gamma$ ranging between 0 and 1 and see the best performance when $\gamma$=0.005 (details in Appendix). Henceforth, unless otherwise mentioned, we will use $\gamma$=0.005 for all our experiments. 
Table \ref{keywordModelPerf} shows the performance of the response generation models. From the table, the KIA for the $no\_kw$ model is negligible, given the one to many nature of open domain dialog. By guiding the model with cues or keywords, the KIA goes up to 67.2\% and this is improved further by the $kw\_loss$ model and is highest at 69.4\%. %{\color{red}We also measure accuracies by finding words similar to the keyword, we use a threshold of 0.7 for the similarity computation.<- WILL PROBABLY REMOVE THE THRESHOLD BASED KIA}  \\
All of the cue/keyword based models outperform the $no\_kw$ model in all of the similarity-based and response quality metrics, except perplexity where the $no\_kw$ model is the best. Adding keyword-loss greatly improves the context coherence and fluency as compared to adding keyword as context information alone. The context coherence is the highest when we use similarity-based keyword loss, which encourages generating sentences with words having similar meaning as the input word. The $kw\_simloss\_glove-1$ and $kw\_simloss\_wordnet-1$ models also show better performance as compared to the $kw\_context$ model. The diversity score and perplexity does drop a bit on the addition of keyword-based losses to the $kw\_context$ model. 
Table \ref{keywordModelPerf} also shows the results for our experiments on using multiple keywords as control during training and also incorporating this into the loss functions. We observe that KIA improves with the $kw\_loss$ models, especially the glove-similarity  based model. More details on these models included in the appendix.

\subsection{Human Evaluation Results}

\paragraph{Task1:}
We collect about 1000 responses for the keywords suggested by the two keyword predictors and also collect 1000 additional human keywords and corresponding responses (resulting in about 2000 responses). We also compute keyword-context-relevance from the data provided by users. From Table \ref{keywordpred}, we see that crowdworkers find keywords generated from generative keyword predictor as more relevant to the conversation context than that generated by the extractive keyword predictor. 
\paragraph{Task2:}%Users interact with our system by reading the conversation context and choosing a keyword and a relevant response, the user can further edit this response or enter a new one. 
Analysing the response choice (human vs model generated) of the turkers, we find that from 121 interactions, 34.7\% of the interactions used model response, and 29.7\% used human response. We also observe that 60 interactions result in edits of the response by the turkers. Out of this, the WER for edits for a human response is 0.45 while WER for edits is lower when a model response is chosen, at 0.39. This indicates that the model response is relevant and captures the content to be conveyed by the user well.
% We capture the turkers time in interaction with our system, with start\_time when the turker finishes reading the provided conversation before making a keyword choice, and ends when a response is chosen. %We also provide a virtual keyboard with a sentence to type and capture the time taken to type the entire sentence. 
% We observe that for the keyword choice and response selection the users took on an average 53.83 seconds (Median = 28.56 seconds).  
Figure \ref{fig:human_task2_eval} shows some statistics on the responses to the questions asked to the user after the above interaction. %The questions try to bring up why the user chose a particular keyword/response pair. 
The plot shows that most people agree/strongly agree that they picked the keyword/response because it seemed relevant to the context or it resonated with the response in their mind. The plot also shows that people did not choose a response because it was short to edit. This analysis shows that our procedure of suggesting keywords followed by relevant responses is the right strategy for building the controllable response generation system. 
% Figure \ref{humanEval} shows the comparison scores for the 4 different model pairs. In terms of Keyword relevance, human evaluation agrees with the automatic metrics and all of our models perform better than the no-keyword model by a very large margin. Although fluency score is seen to drop a bit with the keyword-based models, all of the model pairs have been rated with similar fluency scores by the annotators. All of the model pairs are rated as less generic too, which is very desirable. However, we see that for context relevance metric, in comparison to K-CRG and W-CRG, the no-keyword model does significantly better, while the no-keyword model has similar context relevance as G-CRG. This needs further investigation as a part of our future work. Also consistent with the findings from the automatic evaluation, K-CRG produces more keyword relevant responses as compared to W-CRG. (We did not carry out pairwise evaluations for all possible pairs, we compare all our models with the no-keyword model and pick the best of our models for the fourth comparison). 
\paragraph{Task3:} %In this task, we perform human evaluation to compare the keyword-based models with the model without control. 
Figure \ref{fig:human_task3_eval} shows the scores for the response quality metrics for different model. From human ratings, we observe that the $kw\_loss$ and $kw\_context$ models outperform the model without control, on all metrics significantly. The keyword-based models generate more fluent and relevant responses while at the same time, generating less generic responses compared to the $no\_keyword$ model. We also observe that humans rate $kw\_context$ and $kw\_loss$ models as very comparable, with $kw\_loss$ models being more keyword and context relevant as also established by the automatic evaluations.

\section{Conclusion}
In this paper, we present a novel usage for open domain conversational models - representing differently abled users and enabling them to communicate. In such a use-case, minimizing the need for user intervention is critical, hence the focus of this work has been to develop controllable response generation models that enable fine-grained human control in the form of keyword inputs from the user. We also introduce keyword-based loss functions that encourages the model to generate the keyword or similar words in the response. We show using automatic and human evaluation that these loss functions help in generating more keyword relevant responses. We also extend this to control using multiple keywords, that could further make the models less-restrictive to users. To further improve efficiency and time in interaction, we develop keyword predictors and evaluate them. We show with both automatic and human evaluation that our models outperform the baseline model with no control, at the same time maintaining the response quality. As future work, we plan to broaden the usage of this system to cater to a larger diversity of differently abled population, to help in communication. We are also working with patients to collect feedback and plan to deploy our system as part of a larger AAC system to impact the quality of life of the patients and help the caregivers. Future research direction also involves personalization of these controllable models (both speech and linguistic) that could make these systems more powerful and capable of representing a user.

\bibliography{anthology,custom}
\bibliographystyle{acl_natbib}

\appendix
\label{sec:appendix}
\appendix
\section*{Appendix}

%{\color{red} \textbf{ADD AMT data collection interface}}

\section{Human Evaluation Setup Details}
\begin{figure}[h!]
    \centering
    \includegraphics[width=\columnwidth]{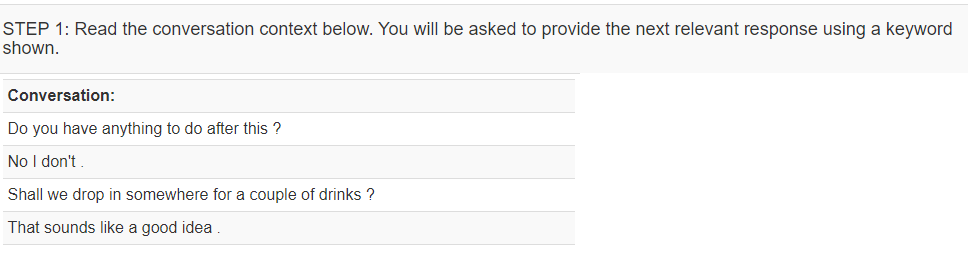}
    \caption{Shows the step 1 for Task 2 on the MTurk study. Here the turkers are presented with the conversation context. }
    \label{fig:step2_1}
\end{figure}

\begin{figure}[h!]
    \centering
    \includegraphics[width=\columnwidth]{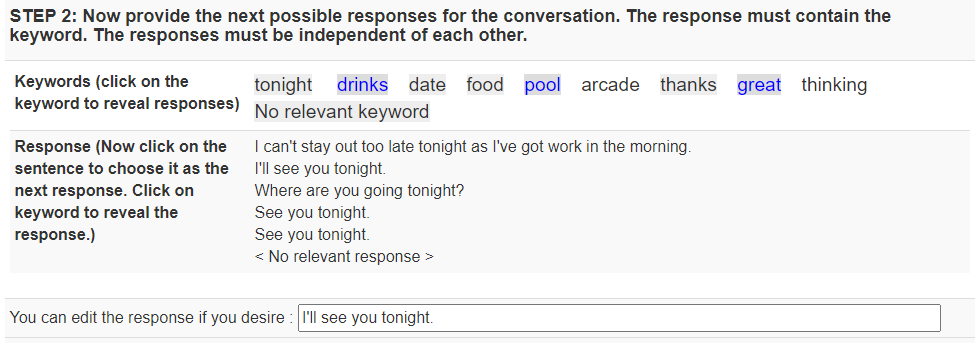}
    \caption{Shows the step 2 for Task 2 on the MTurk study. Here the turkers are shown 9 keywords (generated from keyword predictor models and humans from task 1). Choosing one of them allows them to see the response generated from our models, and human-generated ground truth response, that can be chosen.  }
    \label{fig:step2_2}
\end{figure}

Human evaluation of our system is split into three tasks: task 1 for collecting keywords and corresponding responses from humans. Task 2 involved the crowd workers on Amazon Mechanical Turk interact with our system. We used the keyword suggestions from our extractive and generative keyword predictor models and also the human-generated keywords. We run our controlled response generation pipeline on these keywords to obtain relevant responses. In this task, we first present the turkers with the conversation context as shown in \ref{fig:step2_1}. We also present 9 keyword suggestions - 3 from the extractive keyword predictor, 3 from the generative keyword predictor and 3 keywords generated by humans (from task 1). 
Figure \ref{fig:step2_2} shows this step. Choosing one of these keywords, brings up responses from the human responses generated from Task1, and our controllable response generation model. We use $kw\_loss$ model with $\gamma$=0.005 and diverse beam search to generate the responses.
The users can choose one of the responses and further edit, or enter his/her own response in the box provided. 
% We capture the entire flow and time for this interaction. 

\begin{figure}
    \centering
    \includegraphics[width=\columnwidth]{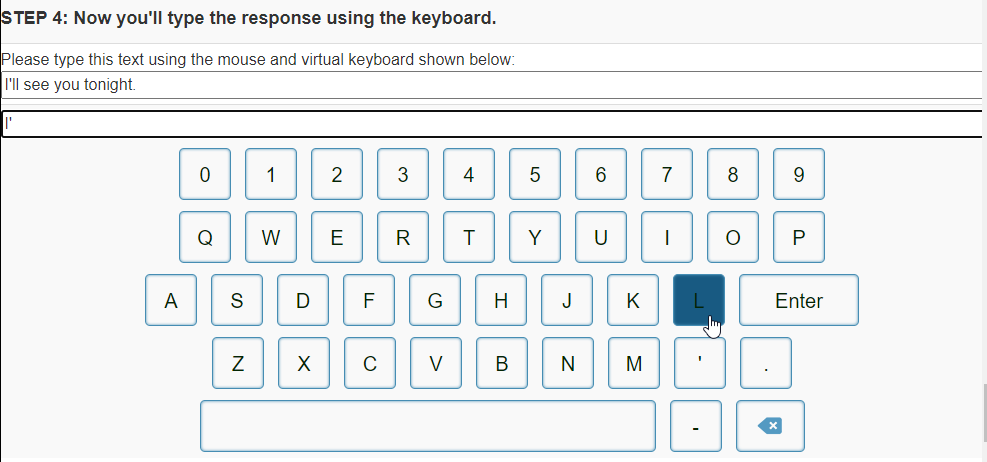}
    \caption{Shows the step 4 for Task 2 on the MTurk study. Step 3 is questionnaire with radio button options which is not shown above. }
    \label{fig:step2_3}
\end{figure}

We then present a questionnaire to the turkers - asking them to answer on a likert scale, some questions about why they chose a particular keyword/responses. At the end, turkers are shown a virtual keyboard as you can see in Figure \ref{fig:step2_3} and asked to type in the response that they chose/edited. Using their physical keyboard is disabled for this part of the task - this is to ensure that the turkers use the virtual keyboard and generate the given text. This data enables us to compare the time it took to complete a single interaction and the time it takes to actually type in the entire response (future work). 

\section{Experiments}

We present the effect of varying the $\gamma$ coefficient in the keyword-based loss models. These results are presented in table \ref{coeffAnalysis}. Please note that when $\gamma = 0$, the model is the $kw\_context$ model. We see from the table that increasing $\gamma$ increases the KIA, which matches our intuition, and reaches close to 75\% when $\gamma =1$. However, we see that this is optimal when $\gamma =0.005$. Similarity metrics such as BLEURT see a drop as we increase $\gamma$ with the lowest at 1. Also, Response Quality deteriorate heavily with context coherence, diversity and fluency metrics. While the higher $\gamma$ tries to increasingly encourage the model to generate the keyword in the sentence, this is at the cost of the overall quality of the response. Hence, in all of the experiments and results reported in the paper, we fix $\gamma =0.005$, unless otherwise specified.

\begin{table*}
\centering
%\resizebox{\textwidth}{!}{
\begin{tabular}{|l|c|c|c|l|c|c|c|c|}
\hline
                      & KWI Accuracy             & \multicolumn{1}{l|}{Similarity}        & BLEURT                          &  \multicolumn{1}{l|}{Context}  & \multicolumn{1}{l|}{Diversity} & \multicolumn{1}{l|}{Fluency} & \multicolumn{1}{l|}{PPL} \\ \hline
coeff=0	&0.672	&0.539&	\textbf{-0.607}	&0.568&	\textbf{1.789}&	0.403&\textbf{	41.752}\\
coeff=0.005	&0.694&	\textbf{0.542}	&-0.609	&0.579	&1.726&	\textbf{0.407}&	43.115\\
coeff=0.01	&0.681&	0.538&	-0.629&	\textbf{0.581}&	1.641&	0.406&	45.749\\
coeff=0.1&	0.690&	0.508&	-0.846&	0.519&	0.888&	0.397&	92.567\\
coeff=1	&\textbf{0.746}&	0.527&	-0.826&	0.468&	0.695&	0.373&	90.070  \\ \hline

\end{tabular}
%}

\caption{\label{coeffAnalysis}Examining the effect of $\gamma$
%$\gamma$ {\color{red} Change this into a figure???}
}
\end{table*}

\section{Sample Model Outputs}
In Table \ref{SampleOuputs}, we present the outputs from the various models - for a given context and keyword. We show the sample outputs from the $no\_kw$, $kw\_context$, $kwloss\_0.005$, $kwloss\_sim\_loss\_glove$ models and the ground truth. We see that the keywords-based models are able to effectively induce the keywords into the generated sentence. 

\begin{table*}
\begin{tabular}{p{2cm} p{7cm}p{7cm}}
\hline \small \textbf{Context} &    
\begin{tabular}{@{}p{7cm}@{}}\small \textbf{Speaker1}: John , I'd like you to meet Charles Brown , our new manager at the airport .  \\ \small \textbf{Speaker2}: Yes , certainly . How can I recognize him ? \\\small \textbf{Speaker1}:  He is short , well-built , and he's got light blond hair . \\ \small\textbf{Speaker2}:    Is there anything else ? \end{tabular}        & 

\begin{tabular}{@{}p{7cm}@{}}\small\textbf{Speaker1}: Please tell me something about your insurance background .  \\ \small \textbf{Speaker2}: I majored in international banking at college so I took many insurance courses . And I worked at an insurance company for my graduation field work . \\ \small\textbf{Speaker1}:  Why do you want to be an insurance agent ? \\. \\\small\textbf{Speaker2}:I care about people and I know the importance of insurance . \\ \small\textbf{Speaker1}: Do you have any work experience in promotion ? \\ \end{tabular}   \\ \hline

\small\textbf{Keyword} &\small\textit{\textbf{mustache}} &\small \textbf{\textit{salesman}} \\ \hline
\small\textbf{NoKeyword} &\small \textit{Yes, he is tall, but he is slim.} &\small \textit{Yes, I have been promoted twice. I have been promoted twice as well.} \\

\small \textbf{kwcontext} &\small\textit{No, he has a very nice mustache.} &\small {\textit{Yes, I have worked as a salesman for a few years.}} \\ %\hline

\small\textbf{KwLoss\_0.005} &\small \textit{He has a rather bushy facial beard and a moustache.}&\small \textit{Yes. I have a sales job as a salesman in a car dealership.}\\
\small\textbf{kwloss\_sim\_glove} & \small\textit{No, that's it. He has a rather light mustache, too.}&\small \textit{Yes, I have been a salesman for two years.}\\
\small\textbf{GroundTruth} &\small \textit{He has a mustache , if I remember correctly , and a light complexion .}& \small\textit{Yes , I do . I have worked as a salesman for P and G Company Ltd .}\\ \\ \\ 
\hline

\hline \small \textbf{Context} &    
\begin{tabular}{@{}p{7cm}@{}}\small \textbf{Speaker1}: Were you in a leadership position when you were a college student ? \\ \small \textbf{Speaker2}:  Yes . I was president of the Student Council of our university . \\\small \textbf{Speaker1}: Did you get any honors or awards at your university ? \\ \small\textbf{Speaker2}:    Yes . I was elected Excellent Party Member for four academic years .\\\small \textbf{Speaker1}: Were you involved in any club activities at your university ?  \\ \small\textbf{Speaker2}:   Yes . I was a member of the Student Poetry Society . I like writing poems in my spare time  \end{tabular}        & 

\begin{tabular}{@{}p{7cm}@{}}\small\textbf{Speaker1}: Do you have a free moment ? \\ \small \textbf{Speaker2}: Sure . What do you need ? \\ \small\textbf{Speaker1}:   I'm having some trouble . How do you use this printer ?  \\. \\\small\textbf{Speaker2}: Well , first you have to turn this switch on . \\ \small\textbf{Speaker1}: Silly me\\ \end{tabular}   \\ \hline

\small\textbf{Keyword} &\small\textit{\textbf{extracurricular}} &\small \textbf{\textit{button}} \\ \hline
\small\textbf{NoKeyword} &\small \textit{What kinds of poems did you write?} &\small \textit{It's not complicated.} \\

\small \textbf{kwcontext} &\small\textit{Did you have any extracurricular activities?} &\small {\textit{Then push the button.}} \\ %\hline

\small\textbf{KwLoss\_0.005} &\small \textit{Were you involved in any extracurricular activities?}&\small \textit{Now press this button.}\\
\small\textbf{kwloss\_sim\_glove} & \small\textit{What was your extracurricular activities?}&\small \textit{Then press the button.}\\
\small\textbf{GroundTruth} &\small \textit{What extracurricular activities did you usually take part in at your college ?}& \small\textit{Now press this button .}\\
\hline

\end{tabular}

\caption{\label{SampleOuputs} Sample conversation contexts and comparison of different model outputs}
\label{modelOutputsSamples}
\end{table*}

\section{Keyword Control with Multiple Inputs}
Table \ref{multikeywordModelPerf} shows the results from our experiments with training the modesl with multiple keywords as control. We see that $kw\_sim\_loss\_wordnet-1$ performs well on several metrics. We plan to look into these models further as part of future work. 

\begin{table*}[tbp]
\centering
\resizebox{\textwidth}{!}{
\begin{tabular}{|lccclcccc|}
\hline
\multicolumn{1}{|l|}{Multiple Keywords}              & \multicolumn{1}{c|}{KIA}         % & \multicolumn{1}{l|}{KIA Threshold}  
& \multicolumn{1}{l|}{Similarity}     & \multicolumn{1}{c|}{BLEURT}           & \multicolumn{1}{l|}{BERT Score}                 & \multicolumn{1}{l|}{Context}        & \multicolumn{1}{l|}{Diversity}      & \multicolumn{1}{l|}{Fluency}        & \multicolumn{1}{l|}{PPL$\downarrow$} \\ \hline
\multicolumn{1}{|l|}{no\_kw} & \multicolumn{1}{c|}{0.041} %         & \multicolumn{1}{c|}{0.070} 
& \multicolumn{1}{c|}{0.271}          & \multicolumn{1}{c|}{-1.035}          & \multicolumn{1}{l|}{0.868/0.836/0.851}          & \multicolumn{1}{c|}{0.541}          & \multicolumn{1}{c|}{1.592}          & \multicolumn{1}{c|}{0.407}          & \textbf{39.098}                   \\ \hline
\multicolumn{1}{|l|}{kw\_context}      & \multicolumn{1}{c|}{0.293}      %    & \multicolumn{1}{c|}{0.122}      
& \multicolumn{1}{c|}{{0.607}} & \multicolumn{1}{c|}{\textbf{-0.499}} & \multicolumn{1}{l|}{\textbf{0.895/0.857/0.875}} & \multicolumn{1}{c|}{0.489}          & \multicolumn{1}{c|}{\textbf{1.396}} & \multicolumn{1}{c|}{0.399}          & {75.300}          \\
\multicolumn{1}{|l|}{kw\_loss}          & \multicolumn{1}{c|}{0.300}      %    & \multicolumn{1}{c|}{0.125}      
& \multicolumn{1}{c|}{0.604}          & \multicolumn{1}{c|}{-0.524}          & \multicolumn{1}{l|}{0.894/0.856/0.874}          & \multicolumn{1}{c|}{\textbf{0.492}} & \multicolumn{1}{c|}{1.354}          & \multicolumn{1}{c|}{0.412}          & 83.971                   \\ \hline
%kw\_sim\_loss                     & \multicolumn{1}{l}{}                & \multicolumn{1}{l}{}                & \multicolumn{1}{l}{}                & \multicolumn{1}{l}{}                 &                                                 & \multicolumn{1}{l}{}                & \multicolumn{1}{l}{}                & \multicolumn{1}{l}{}                & \multicolumn{1}{l|}{}   \\
\multicolumn{1}{|l|}{kw\_sim\_loss\_glove-1} & \multicolumn{1}{c|}{\textbf{0.302}} 
%& \multicolumn{1}{c|}{0.127}          
& \multicolumn{1}{c|}{\textbf{0.610}}          & \multicolumn{1}{c|}{-0.535}          & \multicolumn{1}{l|}{\textbf{0.895/0.857/0.875}} & \multicolumn{1}{c|}{0.487}          & \multicolumn{1}{c|}{1.366}          & \multicolumn{1}{c|}{0.416}          & 84.367                   \\
\multicolumn{1}{|l|}{kw\_sim\_loss\_wordnet-1} & \multicolumn{1}{c|}{0.287}      %    & \multicolumn{1}{c|}{0.120}      
& \multicolumn{1}{c|}{0.600}          & \multicolumn{1}{c|}{-0.525}          & \multicolumn{1}{l|}{0.894/0.856/0.874}          & \multicolumn{1}{c|}{0.488}          & \multicolumn{1}{c|}{1.351}          & \multicolumn{1}{c|}{\textbf{0.417}} & 80.403                   \\ \hline
\multicolumn{1}{|l|}{kw\_sim\_loss\_glove}    & \multicolumn{1}{c|}{0.293}          & %\multicolumn{1}{c|}{\textbf{0.128}} & 
\multicolumn{1}{c|}{0.598}          & \multicolumn{1}{c|}{-0.511}          & \multicolumn{1}{l|}{0.893/0.855/0.873}          & \multicolumn{1}{c|}{0.479}          & \multicolumn{1}{c|}{1.344}          & \multicolumn{1}{c|}{0.412}          & 80.258                   \\
\multicolumn{1}{|l|}{kw\_sim\_loss\_wordnet}     & \multicolumn{1}{c|}{0.300}       %   & \multicolumn{1}{c|}{0.126}       
& \multicolumn{1}{c|}{{0.607}} & \multicolumn{1}{c|}{-0.518}          & \multicolumn{1}{l|}{0.894/0.856/0.875}          & \multicolumn{1}{c|}{0.483}          & \multicolumn{1}{c|}{1.364}          & \multicolumn{1}{c|}{0.416}          & 79.888                   \\ \hline
\end{tabular}
}
\caption{\label{multikeywordModelPerf}Performance of the various controllable models for multiple keyword input ($\gamma$ = 0.005). Label "-1" indicates that we set $sim(k, kw)=1$ in equation \ref{eq3}.}
\end{table*}

% \begin{table*}
% \centering
% \begin{tabular}{ccc}
% \hline \small\textbf{Context} &    
% \begin{tabular}{@{}p{7cm}@{}}\small\textbf{Speaker1}:  Hey, could you help me try and figure out how to get ready for my job interview?  \\  \small\textbf{Speaker2}: The most important thing to do is to make sure you know the company and what services or products it provides. Do you know all about them? \\ \small\textbf{Speaker1}: Yes, I pretty much understand the company.\end{tabular}        &    \begin{tabular}{@{}p{7cm}@{}}\small\textbf{Speaker1}:  Are you going to buy a bicycle? \\  \small\textbf{Speaker2}: I'm thinking about it. How much is it?  \\ \small\textbf{Speaker1}: 350 \$. \\\small\textbf{Speaker2}:Oh, dear. It's too expensive. Can you give me a better price? \\\small \textbf{Speaker1}: Your discount is already included. Normally we charge \$ 400, but now we can give you a special price. This is our sale period.  \end{tabular}   \\ \hline
% \small\textbf{Keyword} &\small\textit{\textbf{research}} & \small\textbf{\textit{cheaper}} \\ \hline
% \small\textbf{Response1} & \small\textit{Do you have any research experience?} &\small \textit{That's much better. I'll take it.} \\
% \small\textbf{Response2} & \small\textit{What kind of research experience do you have?}&\small \textit{That's a lot cheaper than I thought.}\\
% \small\textbf{Response3} &\small \textit{What kind of research do you want to do?}& \small\textit{That's a lot cheaper.}\\
% \hline
% \end{tabular}
% \caption{\label{DiverseBeam} Beam search with Keyword-based model}
% \end{table*}

% 

\end{document}